\documentclass[letterpaper]{article} 
\usepackage[preprint]{aaai2027}
\usepackage[hyphens]{url}  
\usepackage{graphicx} 
\usepackage{natbib}  
\usepackage{caption} 
\usepackage{algorithm}
\usepackage{algorithmic}
\usepackage{amsmath}
\usepackage[table]{xcolor}
\usepackage{multirow}
\usepackage{amssymb}

\usepackage{newfloat}
\usepackage{listings}
\DeclareCaptionStyle{ruled}{labelfont=normalfont,labelsep=colon,strut=off} 
\floatstyle{ruled}
\newfloat{listing}{tb}{lst}{}
\floatname{listing}{Listing}

\usepackage{booktabs}

\title{DECODE: Tackling Representation and Decision Degradation in Continual AI-Generated Image Detection}
\author{
    Zihao Cai\textsuperscript{\rm 1},
    Xinghan Li\textsuperscript{\rm 1},
    Ruiyan Yang\textsuperscript{\rm 1},
    Xue Song\textsuperscript{\rm 1},
    Haijun Shan\textsuperscript{\rm 2},
    Jingjing Chen\textsuperscript{\rm 1}\corresponding
}
\affiliations{
    \textsuperscript{\rm 1}Visual Laboratory of Fudan University, College of Computer Science and Artificial Intelligence,\\
    Fudan University, Shanghai, China\\
    \textsuperscript{\rm 2}CEC GienTech Technology Co., Ltd., Shanghai, China\\
    zhcai25@m.fudan.edu.cn, xinghanli24@m.fudan.edu.cn, 24307140124@m.fudan.edu.cn,\\
    18210860029@fudan.edu.cn, haijun.shan@gientech.com, chenjingjing@fudan.edu.cn
}

\begin{document}

\maketitle

\begin{abstract}
As generative models continue to evolve, AI-generated image detectors must incrementally adapt to emerging generative domains while preserving knowledge acquired from previous ones. This continual learning setting is particularly challenging because forensic traces are often subtle and generator-specific, making detectors highly vulnerable to catastrophic forgetting. 
Existing methods primarily address this problem by stabilizing feature representations, implicitly treating forgetting as a representation-level issue. In this paper, we show that this perspective is incomplete. We demonstrate that even when feature representations remain discriminative, the decision boundary can progressively drift as the classification head is continually optimized on new domains. These two effects jointly give rise to a compound failure mode, termed \textbf{Dual Degradation}. To overcome this challenge, we propose \textbf{DECODE}, a decoupled continual detection framework that jointly mitigates representation- and decision-level forgetting. Specifically, we introduce Subspace Diversity Regularization (SDR) to preserve diverse forensic representations and Closed-Form Decision Alignment (CDA) to recalibrate the shared classification head after each adapter merge without manual hyperparameter tuning. Extensive experiments on 19 generative domains show that DECODE achieves an average accuracy of \textbf{99.36\%} with only \textbf{0.39\%} forgetting, while further generalizing to 11 unseen generators with 95.36\% accuracy.

\end{abstract}

\section{Introduction}

Generative models have advanced rapidly~\cite{karras2021alias,rombach2022high,midjourney}, producing synthetic images that are increasingly difficult to distinguish from real content. The misuse of such technologies poses serious threats, making the detection of AI-generated images an urgent need. While existing detectors achieve strong performance in closed-set settings~\cite{wang2020cnn,li2025towards,yang2026layer}, they struggle to generalize as new generative models continually emerge. This challenge motivates continual learning, which enables detectors to incrementally adapt to new generative domains while retaining previously acquired knowledge.

\begin{figure*}[t]
\centering
\includegraphics[width=\textwidth]{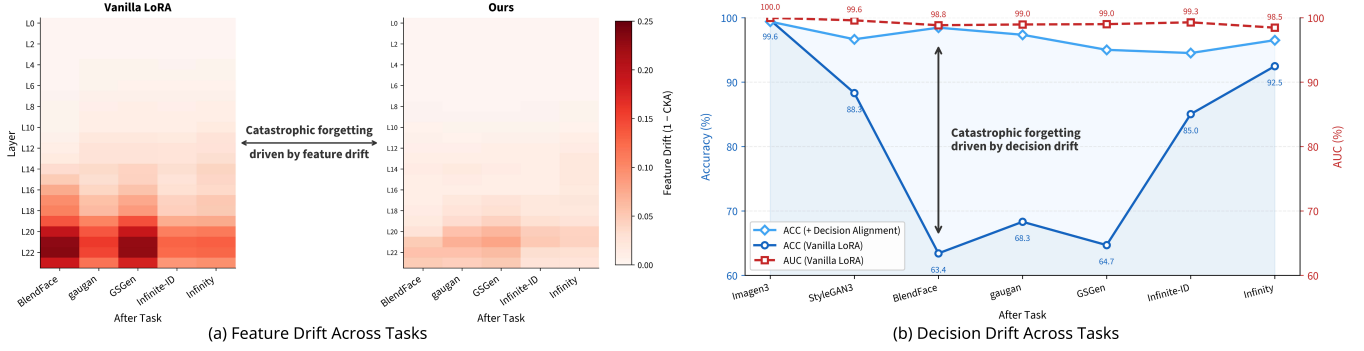}
\caption{Empirical evidence of Dual Degradation. (a) Feature drift on StyleGAN3 as new tasks are learned. Vanilla LoRA shows pronounced drift in deeper layers, while our method largely preserves the learned representations. (b) Decision boundary misalignment on Imagen3. Despite a high AUC, classification accuracy drops as new tasks are introduced. Realigning the decision boundary recovers most of the accuracy drop. Additional empirical evidence is provided in the supplementary material.}
\label{fig1}
\end{figure*}

Compared with standard visual recognition, continual learning for AI-generated image detection is more challenging due to its reliance on subtle, synthesis-specific forensic traces that are easily forgotten during incremental adaptation. To address this, recent methods employ low-rank adaptation (LoRA) to efficiently fine-tune pretrained visual encoders~\cite{zhang2025devfd,hu2025saido,wang2026generalizable}. Despite their different designs, these approaches share a common assumption: catastrophic forgetting primarily arises from degraded feature representations. Consequently, they focus on preserving discriminative forensic features while overlooking the role of the decision boundary. However, our analysis shows that representation degradation alone cannot fully explain forgetting. As illustrated in Figure~\ref{fig1}, sequential LoRA adaptation not only degrades previously learned forensic representations (Figure~\ref{fig1}(a)), but also causes decision boundary drift. Even when feature representations remain highly separable, the drifted decision boundary still leads to substantial performance loss, which can be largely recovered through decision boundary realignment (Figure~\ref{fig1}(b)). These findings reveal that catastrophic forgetting arises from both representation and decision degradation, whereas existing methods address only the former.

In this work, we identify two complementary sources of forgetting: representation degradation and decision degradation, which together form a compound failure mode termed \textbf{Dual Degradation}.
At the representation level, unconstrained LoRA adaptation gradually reduces the diversity of forensic representations by concentrating updates on a few dominant cues, making previously learned features susceptible to overwriting. At the decision level, continual optimization of the shared classification head progressively shifts the decision boundary toward newly introduced domains, degrading performance on earlier ones despite discriminative representations. Since these two degradation pathways originate from different components, i.e., the adapter and the classification head, they require distinct solutions at different stages of continual learning.

To address these distinct degradation pathways, we propose DECODE, a decoupled continual detection framework that tackles representation and decision degradation at their corresponding stages. 
During adapter training, Subspace Diversity Regularization (SDR) explicitly regularizes the LoRA factors by promoting intra-task directional diversity, inter-task subspace separation, and balanced rank utilization. 
This design encourages the adapter to capture complementary forensic cues while reducing cross-task interference, thereby mitigating representation degradation. 
After each adapter merge, Closed-Form Decision Alignment (CDA) recalibrates the shared classification head across all encountered domains via closed-form ridge regression on re-extracted exemplar features, with the regularization strength selected automatically.
This post-merge recalibration realigns the decision boundary as the feature space evolves, thereby mitigating decision degradation without iterative optimization.

The main contributions can be summarized as follows:
\begin{itemize}
    \item We identify \textbf{Dual Degradation} in continual AI-generated image detection, revealing representation and decision degradation as distinct sources of catastrophic forgetting that require separate mitigation.
    \item We propose \textbf{DECODE}, a decoupled continual detection framework that separately mitigates representation and decision degradation. \textbf{SDR} preserves diverse forensic representations during adaptation, while \textbf{CDA} recalibrates the shared classification head across encountered domains through closed-form ridge regression.
    \item Extensive experiments across \textbf{19} heterogeneous generative domains demonstrate that DECODE achieves \textbf{99.36\%} average accuracy with only \textbf{0.39\%} forgetting under continual learning and 95.36\% accuracy on eleven unseen generators.
\end{itemize}

\section{Related Work}

\subsection{AI-Generated Image Detection}

Early studies on AI-generated image detection~\cite{wang2020cnn} showed that detectors trained on a single GAN could generalize to unseen CNN-based generators with appropriate augmentation. Subsequent works have improved cross-generator generalization from various perspectives, including exploiting spatial and frequency artifacts~\cite{tan2024rethinking,tan2024frequency}, enhancing low-level and semantic feature representations~\cite{yan2024sanity,yang2026layer}, and mitigating dataset biases through reconstruction-based detection, artifact-preserving transformations, and bias-free data generation~\cite{wang2023dire,li2025improving,guillaro2025bias}. Meanwhile, pretrained vision-language models have been leveraged through frozen feature representations~\cite{ojha2023towards}, prompt-based adaptation~\cite{tan2025c2p,li2025towards}, and subspace-based feature decomposition~\cite{yan2024orthogonal}, further improving generalization across diverse generators.

Despite these advances, existing methods follow a static training paradigm, assuming a fixed set of generators during training. They cannot continuously adapt to emerging generators while retaining previously learned knowledge, motivating continual learning for AI-generated image detection.

\begin{figure*}[t]
\centering
\includegraphics[width=\textwidth]{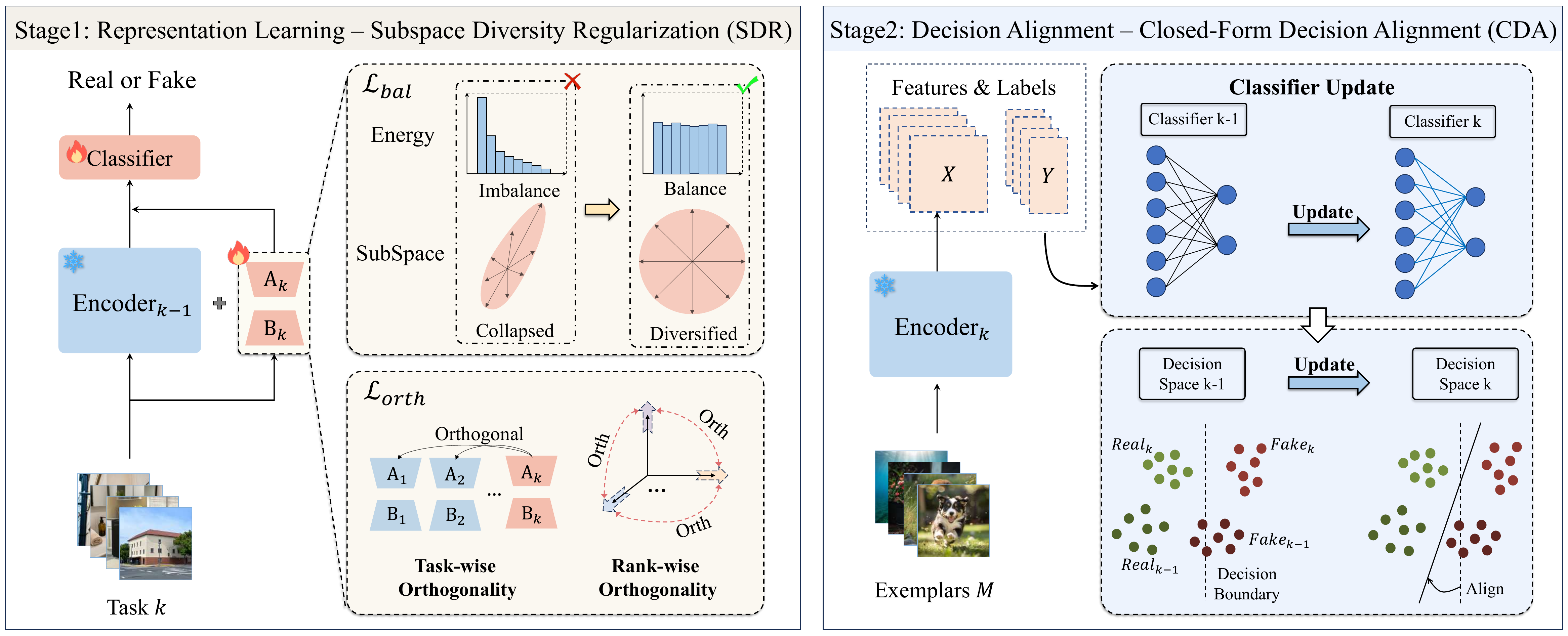}
\caption{Overall framework of DECODE. In Stage~1, SDR regularizes the low-rank adapter to preserve diverse forensic representations and reduce subspace overlap across tasks; the adapter is then merged into the backbone. In Stage~2, CDA recalibrates the classification head via closed-form ridge regression on exemplars from all encountered domains.}
\label{fig:framework}
\end{figure*}

\subsection{AI-Generated Image Detection with Continual Learning}

Continual learning has been widely studied through regularization-based methods that constrain parameter updates~\cite{kirkpatrick2017overcoming,zenke2017continual}, rehearsal-based methods that replay previous examples~\cite{chaudhry2018efficient,chaudhry2019continual}, and architecture-based methods that reduce task interference through parameter isolation~\cite{yoon2017lifelong,serra2018overcoming,yan2021dynamically}. However, these general strategies are not specifically tailored for AI-generated image detection, where subtle and synthesis-specific forensic traces are highly vulnerable to sequential adaptation.

Recent studies have explored continual learning for AI-generated image detection, initially focusing on knowledge preservation through distillation and replay. CoReD~\cite{kim2021cored} adopts knowledge distillation to preserve previous representations, while DFIL~\cite{pan2023dfil} and DMP~\cite{tian2024dynamic} further incorporate replay strategies with sample selection or prototype guidance. More recent approaches improve continual adaptation by explicitly controlling feature evolution, including domain alignment through adapters~\cite{tang2025towards}, latent-space feature isolation~\cite{cheng2025stacking}, task-specific low-rank subspaces~\cite{zhang2025devfd}, and parameter update constraints based on curvature or gradient projection~\cite{wang2026generalizable,zhang2025generalization,hu2025saido}.
Despite these advances, existing methods mainly focus on mitigating representation-level forgetting through feature preservation or constrained adaptation. They overlook decision-level forgetting caused by classification boundary drift as the feature space evolves. In contrast, our work identifies this overlooked degradation pathway and introduces a decoupled solution to address both representation and decision forgetting.

\section{Methodology}

\subsection{Framework Overview}
We consider a continual detection setting with $K$ sequential binary classification tasks $(\mathcal D_1,\ldots,\mathcal D_K)$, where each task introduces a new generative domain and $y\in\{0,1\}$ indicates whether an image is real or generated.

To address the dual degradation in sequential adaptation, DECODE adopts a decoupled two-stage framework, as shown in Figure~\ref{fig:framework}. In \textbf{Stage~1}, we freeze the pretrained visual encoder and jointly optimize a low-rank adapter and the classification head on the current task. SDR regularizes adapter learning to preserve diverse forensic representations and reduce cross-task interference. After training, the adapter is merged into the backbone. In \textbf{Stage~2}, CDA recalibrates the shared classification head using stored exemplars and closed-form ridge regression, yielding a decision boundary aligned across all encountered domains. The updated model is then used for the next task.

\subsection{Subspace Diversity Regularization}

\begin{figure}[!t]
  \centering
  \includegraphics[width=\linewidth]{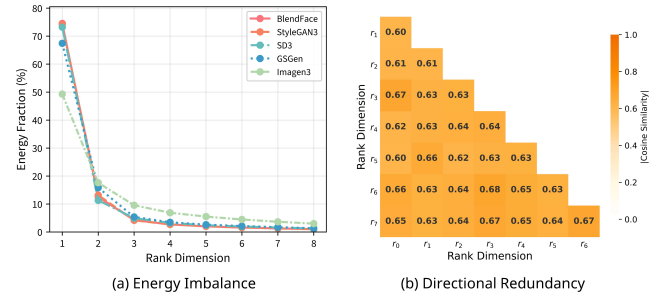}
  \caption{Energy concentration and directional redundancy in vanilla LoRA adapters. (a) Normalized squared Frobenius norms of energy-sorted rank-one update components across five generators. (b) Pairwise absolute Frobenius cosine similarities among rank-one update components in the LoRA adapter trained on the StyleGAN3 task. Further evidence is provided in the supplementary material.}
  \label{fig4}
  \end{figure}

Forensic cues in AI-generated image detection are subtle and heterogeneous across synthesis pipelines. Under unconstrained adaptation, the LoRA update can therefore concentrate on a few dominant cues, reducing representational diversity and making the learned features vulnerable to overwriting by subsequent tasks. We examine this behavior in vanilla LoRA adapters trained for five representative generator tasks. Figure~\ref{fig4} reveals two patterns in the LoRA update geometry: imbalanced rank utilization and directional redundancy. Across the five generators, the highest-energy rank-one component accounts for 50--75\% of the summed component energy, leaving the other components underutilized (Figure~\ref{fig4}(a)). For the StyleGAN3 adapter, pairwise absolute Frobenius cosine similarities among rank-one components consistently exceed 0.60 (Figure~\ref{fig4}(b)). Together, these observations demonstrate uneven rank utilization and substantial directional redundancy in vanilla LoRA. We do not assume a one-to-one correspondence between a rank direction and a manually defined artifact type. Instead, complementary rank directions provide distinct degrees of freedom through which the detector can encode a broader set of synthesis-related cues. SDR therefore regularizes the adapter geometry to promote directional diversity and balanced energy utilization.

\noindent\textbf{Low-Rank Adapter.}
We insert low-rank adapters into the frozen encoder following standard practice. For a weight matrix $W_l$ at layer $l$, the update for task $k$ is parameterized as $\Delta W_l = (\alpha / r)\, B_l^{(k)} A_l^{(k)}$, where $A_l^{(k)} \!\in\! \mathbb{R}^{r \times d_{\text{in}}}$ and $B_l^{(k)} \!\in\! \mathbb{R}^{d_{\text{out}} \times r}$ are low-rank factor matrices, $r$ is the adapter rank, and $\alpha$ is a scaling factor. Each corresponding row of $A_l^{(k)}$ and column of $B_l^{(k)}$ jointly define a rank-one adaptation direction through their outer product, and the $r$ directions together form the adaptation subspace for task $k$. This decomposition depends on the standard LoRA parameterization and is not assumed to be unique for a given update $\Delta W_l$. For each new task, $B_l^{(k)}$ is initialized to zero, ensuring $\Delta W_{l,\text{init}} = \mathbf{0}$ so that training starts from the merged backbone.

\noindent\textbf{Orthogonality Loss.}
Existing orthogonal-subspace methods~\cite{zhang2025devfd} primarily separate task-specific adaptation subspaces to reduce cross-task interference. However, this inter-task objective alone is insufficient, as it leaves correlations among the $r$ rank directions within each task unconstrained. The directional redundancy identified in Figure~\ref{fig4}(b) therefore motivates an orthogonality constraint among these directions. SDR retains the inter-task objective while additionally penalizing intra-task inner products, directly addressing this redundancy. Both constraints act on the incremental LoRA updates rather than the final image representations, allowing related tasks to share features through the merged backbone. Together with the energy balance loss introduced below, SDR therefore regularizes three aspects of the low-rank update: intra-task directional diversity, inter-task subspace separation, and balanced rank utilization. We encode the first two aspects in a unified orthogonality loss.

At layer $l$, we stack the $A$ matrices of all $k$ tasks to form $\widetilde{A}_l \!\in\! \mathbb{R}^{kr \times d_{\text{in}}}$, and similarly form $\widetilde{B}_l \!\in\! \mathbb{R}^{d_{\text{out}} \times kr}$. The global Gram matrices $G_A = \widetilde{A}_l \widetilde{A}_l^\top$ and $G_B = \widetilde{B}_l^\top \widetilde{B}_l$ have size $kr \times kr$. Penalizing their off-diagonal entries yields the orthogonality loss:
\begin{equation}
\mathcal{L}_{\mathrm{orth}} = \sum_l \left( \left\| G_A \odot \bar{I} \right\|_F^2 + \left\| G_B \odot \bar{I} \right\|_F^2 \right),
\label{eq:orth}
\end{equation}
where $\bar{I} = \mathbf{1}_{kr \times kr} - I_{kr}$ masks out the diagonal entries. The block structure of $\bar{I}$ partitions the penalty into two groups. Within each task's diagonal block, off-diagonal entries penalize pairwise inner products among the $r$ directions, encouraging complementary degrees of freedom for encoding synthesis-related cues. Across different tasks, the off-diagonal blocks penalize inner products between task-specific update directions, discouraging subspace overlap and reducing interference during parameter merging.

\noindent\textbf{Energy Balance Loss.}
Beyond directional diversity, balanced energy utilization across rank dimensions is also critical for preventing a few directions from dominating the capacity available for these cues. To address the energy imbalance shown in Figure~\ref{fig4}(a), we penalize the variance of squared factor norms across rank dimensions:
\begin{equation}
\mathcal{L}_{\mathrm{bal}}
=\sum_l\!\left[
\mathrm{Var}_i\!\bigl(\|A_{l,i:}^{(k)}\|_2^2\bigr)
+\mathrm{Var}_i\!\bigl(\|B_{l,:i}^{(k)}\|_2^2\bigr)
\right],
\label{eq:bal}
\end{equation}
where $\mathrm{Var}_i(\cdot)$ denotes the sample variance over the $r$ rank dimensions, and the squared row and column norms measure the energy assigned to the $i$-th direction by the two LoRA factors, respectively. Minimizing these variances encourages all dimensions to contribute equally, allowing the adapter to fully utilize all rank dimensions.

\noindent\textbf{Training Objective.}
The complete objective for task $k$ combines the classification loss with the two regularization terms:
\begin{equation}
\mathcal{L}_{\mathrm{total}} = \mathcal{L}_{\mathrm{CE}} + \lambda_{\mathrm{orth}} \, \mathcal{L}_{\mathrm{orth}} + \lambda_{\mathrm{bal}} \, \mathcal{L}_{\mathrm{bal}},
\label{eq:total}
\end{equation}
where $\mathcal{L}_{\mathrm{CE}}$ denotes the softmax cross-entropy loss.

\noindent\textbf{Parameter Merging.}
After training on task $k$, the adapter is merged into the backbone: $W_l^{(k)} = W_l^{(k-1)} + (\alpha/r)\, B_l^{(k)} A_l^{(k)}$. The factor matrices are then retained as frozen tensors for computing $\mathcal{L}_{\mathrm{orth}}$ in subsequent tasks. Unlike multi-expert architectures~\cite{hu2025saido,zhang2025devfd}, which maintain separate adapters for each task, our merging strategy yields a single unified backbone. This keeps inference cost constant regardless of the number of tasks and requires no task identification or routing at test time.

\subsection{Closed-Form Decision Alignment}
Training the shared classification head on the current domain biases the decision boundary toward recently observed generators, while adapter merging continuously changes the underlying feature space. To mitigate this decision drift, CDA recalibrates the shared classification head after each adapter merge using all stored exemplars and closed-form ridge regression.
This repeated alignment allows the classifier to adapt to the evolving feature space, while automatic regularization selection eliminates manual hyperparameter tuning. Since ridge regression is strictly convex, the optimal solution is uniquely determined by the data and can be obtained through a single closed-form computation. We next introduce the exemplar memory and recalibration procedure.

\noindent\textbf{Exemplar Memory.}
After each task, we update a compact memory $\mathcal{M}$ by randomly storing $m$ images per class. Random sampling is adopted because the backbone representation changes after adapter merging, making feature-dependent selection strategies potentially unstable across tasks. Since CDA re-extracts features from all stored exemplars using the latest backbone, random sampling provides sufficient coverage of the input distribution.

\noindent\textbf{Closed-Form Recalibration.}
After merging the adapter for task $k$, we extract features for all stored exemplars using the updated backbone to obtain $X \!\in\! \mathbb{R}^{N \times d}$. Let $\tilde{X} = [X, \mathbf{1}] \!\in\! \mathbb{R}^{N \times (d+1)}$ denote the augmented feature matrix and $Y \!\in\! \mathbb{R}^{N \times 2}$ the one-hot label matrix. The recalibrated head parameters are given by the closed-form ridge regression solution:
\begin{equation}
\tilde{W}^* = \left( \tilde{X}^\top \tilde{X} + \lambda \tilde{I} \right)^{-1} \tilde{X}^\top Y,
\label{eq:ada}
\end{equation}
where $\tilde{I}$ is a diagonal regularization matrix whose last entry is zero so that the bias remains unregularized. The strict convexity of the objective ensures a unique, globally optimal solution that is fully determined by the data and $\lambda$. In practice, CDA requires solving only a single linear system, avoiding iterative optimization entirely.

\noindent\textbf{Automatic $\lambda$ Selection.}
The optimal regularization strength $\lambda$ varies as the feature distribution evolves across tasks. A fixed validation set is unreliable in this setting, because the data distribution shifts after each task. We instead select $\lambda$ automatically through Leave-One-Out Cross-Validation (LOO-CV), which estimates the prediction error directly on the current exemplar memory. To account for the unregularized bias, we center the features as $X_c=X-\mathbf{1}\bar{x}^{\top}$, where $\bar{x}$ is the empirical feature mean, and compute a one-time SVD $X_c=USV^\top$. For each candidate $\lambda$, the diagonal of the corresponding hat matrix and the LOO residuals are computed in closed form:
\begin{equation}
\begin{aligned}
H_{ii}(\lambda) &= \frac{1}{N}+\textstyle\sum_j U_{ij}^2 \frac{s_j^2}{s_j^2+\lambda}, \\
e_i(\lambda) &= \frac{Y_{i,:}-\hat{Y}_{i,:}(\lambda)}{1-H_{ii}(\lambda)},
\end{aligned}
\label{eq:loo}
\end{equation}
where $s_j$ is the $j$-th singular value of $X_c$, $\hat{Y}_{i,:}(\lambda)$ is the fitted label vector for exemplar $i$, and the term $1/N$ accounts for the unregularized bias. The optimal strength is selected as $\lambda^* = \arg\min_\lambda \frac{1}{N}\sum_i \|e_i(\lambda)\|^2$. This makes CDA fully automatic across the entire learning sequence, requiring no manual tuning or held-out validation data.

\begin{table*}[t!]
  \centering

  \fontsize{9pt}{10.8pt}\selectfont
  \setlength{\tabcolsep}{3.2pt}
  \begin{tabular}{l c c c c c c c c c c c c c c c c}
    \toprule
    \multirow{2}{*}{Method}
    & 1-SD3
    & \multicolumn{2}{c}{2-Imagen3}
    & \multicolumn{2}{c}{3-StyleGAN3}
    & \multicolumn{2}{c}{4-BlendFace}
    & \multicolumn{2}{c}{5-GauGAN}
    & \multicolumn{2}{c}{6-GSGen}
    & \multicolumn{2}{c}{7-Infinite-ID}
    & \multicolumn{2}{c}{8-Infinity}
    & \multirow{2}{*}{\begin{tabular}[c]{@{}c@{}}New\\ACC\end{tabular}} \\
    \cmidrule(l){2-16}
    & AA & AA & AF & AA & AF & AA & AF & AA & AF & AA & AF & AA & AF & AA & AF \\
    \midrule
    \rowcolor{gray!7}
    \multicolumn{17}{l}{\textbf{Non-Continual Methods}} \\

    NPR~\citeyearpar{tan2024rethinking}
    & 89.50 & 92.29 & \textbf{-2.06}
    & 63.07 & 41.22
    & 62.35 & 49.59
    & 60.36 & 41.80
    & 59.60 & 42.24
    & 57.31 & 44.26
    & 68.76 & 30.33
    & 95.05 \\

    SAFE~\citeyearpar{li2025improving}
    & 80.88 & 74.86 & 25.26
    & 50.60 & 49.99
    & 59.98 & 36.75
    & 65.07 & 30.24
    & 59.99 & 37.22
    & 62.72 & 34.42
    & 80.16 & 14.85
    & 93.15 \\

    Effort~\citeyearpar{yan2024orthogonal}
    & 99.52 & \underline{99.32} & 0.06
    & \underline{97.35} & \underline{3.16}
    & 88.50 & 14.76
    & 87.69 & 14.96
    & \underline{96.43} & \underline{3.94}
    & \underline{97.95} & \underline{2.02}
    & \underline{98.68} & \underline{1.18}
    & \underline{99.71} \\

    IAPL~\citeyearpar{li2025towards}
    & 99.50 & 99.12 & 0.04
    & 95.66 & 4.55
    & 76.52 & 29.87
    & 86.41 & 15.87
    & 84.43 & 17.79
    & 87.79 & 13.36
    & 91.49 & 8.95
    & 99.32 \\

    LTD~\citeyearpar{yang2026layer}
    & 98.92 & 98.97 & 0.08
    & 93.65 & 3.25
    & \underline{95.25} & \textbf{1.29}
    & \underline{96.60} & \textbf{0.14}
    & 89.66 & 9.52
    & 95.07 & 2.43
    & 94.75 & 3.15
    & 96.81 \\

    \rowcolor{gray!7}
    \multicolumn{17}{l}{\textbf{Continual Methods without Replay}} \\

    DevFD~\citeyearpar{zhang2025devfd}
    & 98.92 & 99.16 & \underline{-0.56}
    & 86.61 & 18.25
    & 64.42 & 46.20
    & 77.35 & 27.35
    & 68.79 & 36.68
    & 82.63 & 19.58
    & 85.03 & 16.52
    & 99.41 \\

    SAIDO~\citeyearpar{hu2025saido}
    & \underline{99.54} & 98.53 & 0.28
    & 92.44 & 3.78
    & 91.13 & 6.02
    & 92.22 & 4.81
    & 91.02 & 6.80
    & 93.96 & 3.05
    & 93.79 & 2.97
    & 96.18 \\

    \rowcolor{gray!20}
    DECODE (Ours)
    & \textbf{99.58} & \textbf{99.59} & -0.18
    & \textbf{97.82} & \textbf{2.50}
    & \textbf{96.07} & \underline{4.73}
    & \textbf{97.16} & \underline{3.14}
    & \textbf{96.64} & \textbf{3.70}
    & \textbf{98.23} & \textbf{1.76}
    & \textbf{99.14} & \textbf{0.70}
    & \textbf{99.73} \\

    \midrule
    \rowcolor{gray!7}
    \multicolumn{17}{l}{\textbf{Continual Methods with Replay}} \\

    DFIL~\citeyearpar{pan2023dfil}
    & \underline{98.84} & \underline{98.36} & \underline{0.30}
    & 89.64 & 8.23
    & 83.09 & 17.61
    & 81.45 & 18.62
    & 82.79 & 16.95
    & 80.99 & 18.90
    & 88.11 & 10.32
    & \underline{97.17} \\

    SUR-LID~\citeyearpar{cheng2025stacking}
    & 96.57 & 96.51 & 0.66
    & 67.05 & 39.56
    & 62.74 & 43.03
    & 67.72 & 34.15
    & 58.75 & 44.54
    & 62.79 & 39.24
    & 61.19 & 40.61
    & 96.72 \\

    Tang et al.~\citeyearpar{tang2025towards}
    & 98.22 & 94.60 & 0.90
    & \underline{91.33} & \underline{1.30}
    & \underline{92.19} & \underline{1.90}
    & \underline{91.73} & \underline{1.14}
    & \underline{90.78} & \underline{3.26}
    & \underline{92.07} & \underline{1.93}
    & \underline{91.30} & \underline{1.78}
    & 92.86 \\

    \rowcolor{gray!20}
    DECODE (Ours)
    & \textbf{99.58} & \textbf{99.64} & \textbf{-0.14}
    & \textbf{98.83} & \textbf{0.84}
    & \textbf{98.14} & \textbf{1.81}
    & \textbf{98.80} & \textbf{0.96}
    & \textbf{98.93} & \textbf{0.84}
    & \textbf{99.34} & \textbf{0.39}
    & \textbf{99.36} & \textbf{0.39}
    & \textbf{99.65} \\

    \bottomrule
  \end{tabular}
  
  \caption{Comparison with state-of-the-art methods under Protocol~1 (\%). Best results are in bold and second-best are underlined.}
  \label{tab:protocol1}
\end{table*}

\section{Experiments}

\begin{table*}[t!]
  \centering

  {
  \fontsize{9pt}{10.8pt}\selectfont

  \setlength{\tabcolsep}{2.0pt}

  \begin{tabular}{l c c c c c c c c c c c c}
    \toprule
    \multirow{2}{*}{Method}
    & \multicolumn{11}{c}{Out-of-Distribution (OOD) Generators}
    & \multirow{2}{*}{Avg.} \\
    \cmidrule(lr){2-12}
    & BigGAN
    & R3GAN
    & InSwap
    & Midjourney
    & DALLE3
    & GaussCtrl
    & \begin{tabular}[c]{@{}c@{}}FLUX.2-\\dev\end{tabular}
    & \begin{tabular}[c]{@{}c@{}}GPT-Image-\\1.5\end{tabular}
    & \begin{tabular}[c]{@{}c@{}}Nano-\\Banana\end{tabular}
    & \begin{tabular}[c]{@{}c@{}}Seedream\\4.5\end{tabular}
    & SORA
    & \\
    \midrule

    NPR~\citeyearpar{tan2024rethinking}
    & 87.39 & 43.24 & 66.57 & 80.55 & 74.71
    & 85.10 & 88.06 & 84.42 & 86.25 & 89.25
    & 83.30 & 78.99 \\

    SAFE~\citeyearpar{li2025improving}
    & 70.63 & 79.52 & 78.06 & 87.85 & 47.58
    & 51.90 & 59.77 & 90.30 & 93.40 & 50.00
    & 50.70 & 69.06 \\

    Effort~\citeyearpar{yan2024orthogonal}
    & 85.84 & \underline{96.48} & \underline{99.61}
    & 92.65 & 72.44 & \underline{98.70}
    & 78.54 & 86.27 & 93.58 & 52.00
    & 60.38 & 83.32 \\

    IAPL~\citeyearpar{li2025towards}
    & \underline{92.48} & 59.38 & 79.29
    & 90.10 & 86.36 & 98.47
    & \underline{93.97} & \underline{90.91}
    & \underline{95.85} & \underline{91.27}
    & \underline{90.85} & 88.08 \\

    LTD~\citeyearpar{yang2026layer}
    & 92.41 & 67.24 & 94.76 & 91.80
    & \underline{94.00} & 95.88
    & 92.42 & 90.18 & 90.17 & 86.25
    & 84.67 & \underline{89.07} \\

    \midrule

    SUR-LID~\citeyearpar{cheng2025stacking}
    & 53.95 & 50.26 & 52.67 & 60.48 & 72.21
    & 49.98 & 61.06 & 54.79 & 57.73 & 55.75
    & 57.65 & 56.96 \\

    DFIL~\citeyearpar{pan2023dfil}
    & 69.49 & 53.42 & 93.82 & 82.25 & 83.16
    & 73.05 & 81.10 & 67.97 & 82.15 & 71.75
    & 79.75 & 72.40 \\

    Tang et al.~\citeyearpar{tang2025towards}
    & 76.57 & 50.50 & 96.59 & 85.05 & 85.51
    & 90.40 & 77.64 & 80.74 & 81.01 & 63.25
    & 76.70 & 78.54 \\

    DevFD~\citeyearpar{zhang2025devfd}
    & 75.43 & 50.32 & 83.47 & 83.80 & 88.33
    & 94.00 & 92.19 & 72.71 & 92.26 & 74.75
    & 75.82 & 80.28 \\

    SAIDO~\citeyearpar{hu2025saido}
    & 82.45 & 93.56 & 89.51
    & \underline{92.90} & 85.86 & 88.35
    & 93.42 & 84.34 & 86.35 & 81.75
    & 77.45 & 86.90 \\

    \rowcolor{gray!20}
    DECODE (Ours)
    & \textbf{93.26}
    & \textbf{98.26}
    & \textbf{99.80}
    & \textbf{93.50}
    & \textbf{96.63}
    & \textbf{99.00}
    & \textbf{94.98}
    & \textbf{92.00}
    & \textbf{97.59}
    & \textbf{92.50}
    & \textbf{91.47}
    & \textbf{95.36} \\

    \bottomrule
  \end{tabular}
  }

  \caption{Open-world generalization performance under Protocol~2 (\%). Best results are in bold and second-best are underlined.}
  \label{tab:protocol2}
\end{table*}

\subsection{Experimental Setup}

\noindent\textbf{Datasets.}
Our evaluation spans 19 generative models covering diverse synthesis paradigms, including diffusion models, GANs, face manipulation methods, 3D Gaussian Splatting (3DGS), and autoregressive visual generators. We collect data from several public benchmarks. AIGIBench~\cite{li2025artificial} provides paired real and generated images for nine generators: SD3~\cite{SD3}, Imagen3~\cite{Imagen}, StyleGAN3~\cite{karras2021alias}, BlendFace~\cite{shiohara2023blendface}, Infinite-ID~\cite{wu2024infinite}, R3GAN~\cite{huang2024gan}, InSwap~\cite{wang2023inswap}, Midjourney~\cite{midjourney}, and DALLE3~\cite{openai2023dalle3}. GauGAN~\cite{park2019semantic} and BigGAN~\cite{brock2018large} are obtained from CDDB~\cite{li2023continual}, Infinity~\cite{han2025infinity} from ARForensics~\cite{zhang2025d3qe}, and GSGen~\cite{chen2024text} and SORA~\cite{openai2024sora} from NeuroRenderedFake~\cite{dongneurorenderedfake}. For FLUX.2-dev~\cite{bfl2025flux2}, GPT-Image-1.5~\cite{openai2025gptimage}, Nano-Banana~\cite{sharon2025nanobanana}, and Seedream4.5~\cite{bytedance2025seedream}, we use generated images from T2I-CoReBench~\cite{li2025easier} paired with real images from COCO~\cite{lin2014microsoft}. For GaussCtrl~\cite{wu2024gaussctrl}, we generate images with the public model using ImageNet~\cite{deng2009imagenet} scenes.

\noindent
\textbf{Evaluation Protocols.} We establish two evaluation protocols to assess continual detection performance and open-world generalization, respectively. Results under additional task orders are reported in the supplementary material.

\begin{itemize}

\item \textbf{Protocol~1} arranges eight generators as sequential continual learning tasks: SD3 $\to$ Imagen3 $\to$ StyleGAN3 $\to$ BlendFace $\to$ GauGAN $\to$ GSGen $\to$ Infinite-ID $\to$ Infinity. This ordering is deliberately designed as a heterogeneous stress test that interleaves distinct synthesis paradigms, requiring the detector to handle abrupt shifts in artifact distributions without forgetting prior ones.
\item \textbf{Protocol~2} evaluates open-world generalization by testing the final model trained under Protocol~1, without further adaptation, on eleven generators entirely unseen during training: BigGAN, R3GAN, InSwap, Midjourney, DALLE3, GaussCtrl, FLUX.2-dev, GPT-Image-1.5, Nano-Banana, Seedream4.5, and SORA, ranging from classical GANs to recent generative models.
\end{itemize}

\noindent
\textbf{Metrics.} We define a test accuracy matrix $R \in \mathbb{R}^{K \times K}$, where $R_{i,j}$ records the accuracy on task $j$ after training through task $i$. After training on task $k$, we report \textbf{Average Accuracy} (AA) $= \frac{1}{k}\sum_{j=1}^{k} R_{k,j}$, which measures the mean detection accuracy across all $k$ encountered tasks. For $k>1$, we report \textbf{Average Forgetting} (AF) $= \frac{1}{k-1}\sum_{j=1}^{k-1} (R_j^* - R_{k,j})$, where $R_j^* \!=\! \max_{j \leq t < k} R_{t,j}$; this metric measures the mean accuracy decline relative to each task's historical best. Over the full $K$-task sequence, we report \textbf{New Accuracy} (New ACC) $= \frac{1}{K}\sum_{i=1}^{K} R_{i,i}$, which averages the accuracy on each task immediately after it is learned. AF directly measures the stability of the detector against forgetting, while New ACC reflects its plasticity.

\noindent
\textbf{Compared Methods.} We compare five non-continual detectors---NPR~\cite{tan2024rethinking}, SAFE~\cite{li2025improving}, Effort~\cite{yan2024orthogonal}, IAPL~\cite{li2025towards}, and LTD~\cite{yang2026layer}---and five continual learning methods: DevFD~\cite{zhang2025devfd}, DFIL~\cite{pan2023dfil}, SUR-LID~\cite{cheng2025stacking}, \citet{tang2025towards}, and SAIDO~\cite{hu2025saido}. At step $k$, each non-continual detector is sequentially fine-tuned only on the current-task data $\mathcal{D}_k$. For continual methods, we distinguish replay-free and replay-based settings. We evaluate all methods using their official implementations with identical dataset splits and evaluation protocols. Each experiment is repeated three times, and we report the mean results.

\noindent
\textbf{Implementation Details.} We use CLIP ViT-L/14 with rank-$8$ LoRA adapters and $224\times224$ inputs. Training uses Adam with a learning rate of $10^{-3}$, batch size 16, and 10 epochs per task; $\lambda_{\mathrm{orth}}=\lambda_{\mathrm{bal}}=1$ and the exemplar memory contains $m=50$ images per class. CDA selects $\lambda$ automatically via LOO-CV (Eq.~(\ref{eq:loo})). For the replay-free variant, which has no exemplar memory, we disable CDA, apply only SDR during adapter training, and freeze the classification head after the first task to mitigate decision degradation.

\subsection{Performance on Continual Learning Tasks}

For a fair comparison, Table~\ref{tab:protocol1} compares the replay-free DECODE variant with both non-continual methods and continual methods without replay, whereas the full DECODE model, which uses an exemplar memory for CDA, is compared only with continual methods using replay. In the replay-free setting, DECODE maintains AA above 96\% throughout the sequence and finishes with 99.14\% AA and 0.70\% AF. It outperforms both the strongest replay-free continual baseline, SAIDO (93.79\% AA and 2.97\% AF), and the best non-continual method, Effort (98.68\% AA and 1.18\% AF). With replay, the full DECODE model maintains AA above 98\% and reaches 99.36\% AA and 0.39\% AF after the final task, compared with 91.30\% AA and 1.78\% AF for the strongest replay-based baseline, Tang et al. Both DECODE variants achieve a New ACC of at least 99.65\%, indicating that their stability does not come at the expense of plasticity.

The AA trajectory across tasks further reveals how the performance gap evolves as the task sequence becomes more heterogeneous. On the first two diffusion-based tasks, DECODE and SAIDO differ by at most 1.06 percentage points. When StyleGAN3 is introduced at task~3, SAIDO drops from 98.53\% to 92.44\% AA, whereas DECODE retains 97.82\%. DECODE subsequently leads SAIDO by 4.27--5.62 points across the remaining tasks, which span face manipulation, scene generation, and 3DGS. These results demonstrate that DECODE adapts effectively across distinct synthesis mechanisms while retaining performance on earlier domains.

\begin{figure*}[t!]
  \centering
  \includegraphics[width=\textwidth]{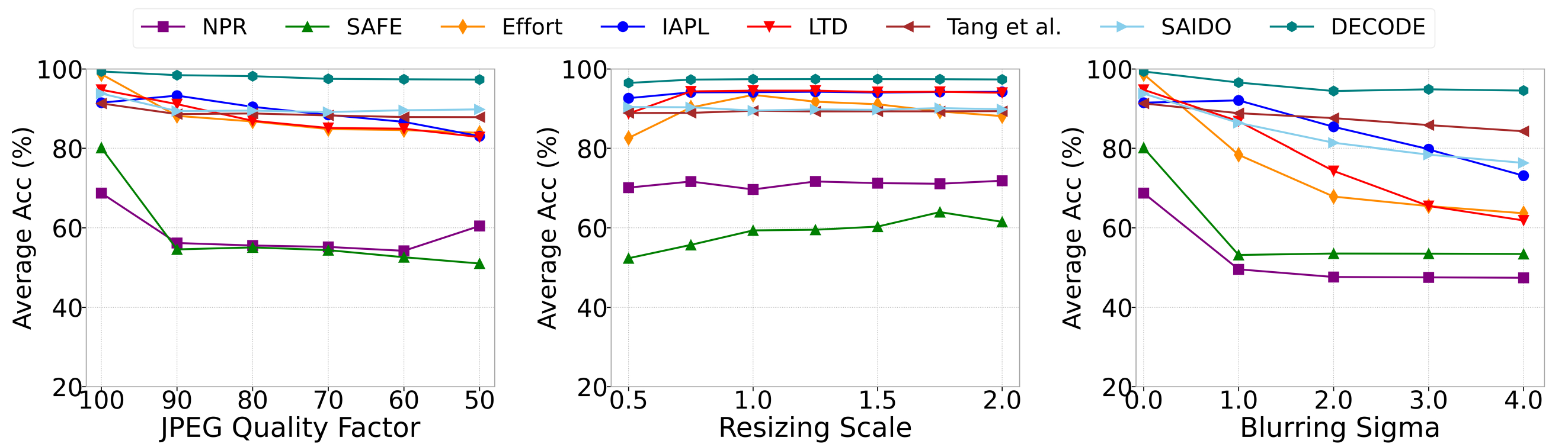}
  \caption{Robustness under JPEG compression, resizing, and Gaussian blur on the Protocol~1 test sets.}
  \label{fig:robustness}
\end{figure*}

\begin{table*}[t!]
  \centering

  \setlength{\tabcolsep}{2.1pt}
  \fontsize{9pt}{10.5pt}\selectfont
  \begin{tabular}{l c c c c c c c c c c c c c c c}
    \toprule
    \multirow{2}{*}{Method} & 1-SD3 & \multicolumn{2}{c}{2-Imagen3} & \multicolumn{2}{c}{3-StyleGAN3} & \multicolumn{2}{c}{4-BlendFace} & \multicolumn{2}{c}{5-GauGAN} & \multicolumn{2}{c}{6-GSGen} & \multicolumn{2}{c}{7-Infinite-ID} & \multicolumn{2}{c}{8-Infinity}\\
    \cmidrule(lr){2-16}
     & AA & AA & AF & AA & AF & AA & AF & AA & AF & AA & AF & AA & AF & AA & AF\\
    \midrule
    Vanilla LoRA                                & \textbf{99.76} & \underline{99.67} & -0.02 & 90.43 & 13.62  & 69.46 & 40.23 & 80.18 & 24.39 & 81.80 & 21.54 & 95.89 & 4.52 & 96.91 & 3.30\\
    \hspace{1em} + $\mathcal{L}_{\mathrm{orth}}^{\mathrm{inter}}$  & 99.58 & 99.56 & -0.12 & 93.23 & 9.46  & 78.75 & 27.85 & 83.69 & 20.02 & 88.34 & 13.69 & 98.08 & 1.97 & 98.46 & 1.52\\
    \hspace{1em} + $\mathcal{L}_{\mathrm{orth}}$         & 99.58 & 99.64 & -0.14 & 97.92 & 2.50  & 90.46 & 12.30 & 92.25 & 9.35 & 93.80 & 7.17 & 98.49 & 1.51 & 98.89 & 1.02\\
    \hspace{1em} + $\mathcal{L}_{\mathrm{bal}}$          & \underline{99.74} & \textbf{99.68} & 0.00 & 93.47 & 9.35  & 74.38 & 33.86 & 83.08 & 20.89 & 83.53 & 19.56 & 95.70 & 4.82 & 96.44 & 3.89\\
    \hspace{1em} + SDR                          & 99.58 & 99.62 & -0.20 & 97.89 & 2.52  & 91.55 & 10.84 & 93.43 & 7.87 & 94.26 & 6.61 & 98.19 & 1.86 & 99.10 & 0.81 \\
    \hspace{1em} + SDR + GD                     & 99.54 & 99.61 & \underline{-0.22} & 98.09 & 2.29  & 94.39 & 7.09 & 96.24 & 4.38 & \underline{96.95} & \underline{3.40} & \underline{98.72} & \underline{1.25} & \underline{99.11} & \underline{0.80} \\
    \hspace{1em} + CDA                          & 99.58 & 99.56 & \textbf{-0.24} & \underline{98.52} & \underline{1.65}  & \underline{96.41} & \underline{4.28} & \underline{98.28} & \underline{1.70} & 96.45 & 3.85 & 97.01 & 3.10 & 97.76 & 2.23 \\ 
    \rowcolor{gray!20} \hspace{1em} + SDR + CDA (Ours)  & 99.58 & 99.64 & -0.14 & \textbf{98.83} & \textbf{0.84} & \textbf{98.14} & \textbf{1.81} & \textbf{98.80} & \textbf{0.96} & \textbf{98.93} & \textbf{0.84} & \textbf{99.34} & \textbf{0.39} & \textbf{99.36} & \textbf{0.39}\\
    \bottomrule
  \end{tabular}
  \caption{Ablation study of components (\%).
$\mathcal{L}_{\mathrm{orth}}^{\mathrm{inter}}$ retains only the inter-task penalty, whereas $\mathcal{L}_{\mathrm{orth}}$ additionally penalizes intra-task inner products. GD denotes gradient-descent-based decision alignment using the same exemplar memory as CDA.}
  \label{tab:ablation}
\end{table*}

\subsection{Open-World Generalization Performance}

After completing the Protocol~1 sequence, we directly evaluate the final model of each method under Protocol~2 on eleven generators entirely unseen during training, without further adaptation. As shown in Table~\ref{tab:protocol2}, DECODE achieves the highest average accuracy of 95.36\%, outperforming the second-best method, LTD (89.07\%), by 6.29 percentage points. This advantage holds across all eleven unseen generators: DECODE performs best on every generator and maintains accuracy above 91\% in all cases.

These gains across GANs, face manipulation, 3DGS, and recent text-to-image models demonstrate effective transfer across synthesis paradigms. We attribute DECODE's strong open-world generalization primarily to SDR, which preserves diverse forensic representations. Ablation results in the supplementary material support this interpretation: SDR alone achieves 95.15\% accuracy under Protocol~2, compared with 90.25\% for CDA alone and 95.36\% for full DECODE.

\subsection{Robustness to Perturbations}

Figure~\ref{fig:robustness} shows three robustness evaluations on the Protocol~1 test sets for representative methods. DECODE exhibits strong robustness across JPEG compression, resizing, and Gaussian blur, outperforming the second-best method by 7.52, 3.13, and 10.26 percentage points under JPEG 50, RESIZE 2.0, and BLUR 4.0, respectively. We attribute this consistent robustness to its decoupled design, which jointly preserves diverse forensic representations and aligns the decision boundary under post-processing shifts.

\subsection{Ablation Study}

Table~\ref{tab:ablation} isolates the contributions of SDR and CDA under Protocol~1. Compared with $\mathcal{L}_{\mathrm{orth}}^{\mathrm{inter}}$, the full $\mathcal{L}_{\mathrm{orth}}$ improves AA/AF from 78.75\%/27.85\% to 90.46\%/12.30\% after learning BlendFace and from 98.46\%/1.52\% to 98.89\%/1.02\% after the final task. These results show that inter-task subspace separation alone is insufficient and that regularizing intra-task directional correlations provides an additional gain. Adding $\mathcal{L}_{\mathrm{bal}}$ to the full $\mathcal{L}_{\mathrm{orth}}$ further improves the final AA/AF to 99.10\%/0.81\%, indicating that balanced rank utilization complements orthogonality regularization.

CDA is effective both independently and when combined with SDR. Applied to Vanilla LoRA, CDA reaches 97.76\% AA and 2.23\% AF after the final task. To determine whether its gains simply result from updating the classification head on replay data, we replace CDA with GD while retaining the same exemplars. From StyleGAN3 onward, GD consistently improves upon SDR, but CDA achieves higher AA and lower AF at every task. After learning BlendFace, SDR+CDA reaches 98.14\%/1.81\%, compared with 94.39\%/7.09\% for SDR+GD. This advantage persists through the final task, where SDR+CDA achieves 99.36\%/0.39\% versus 99.11\%/0.80\% for SDR+GD. Together, CDA's gain over Vanilla LoRA and consistent advantage over GD demonstrate that its closed-form post-merge recalibration effectively mitigates decision degradation as the feature space evolves.

\section{Conclusion}
This work identifies Dual Degradation in continual AI-generated image detection, where representation drift and decision boundary drift jointly cause forgetting. Accordingly, we propose DECODE, a decoupled continual detection framework that addresses these two degradation pathways separately. SDR regularizes the low-rank update geometry to preserve diverse forensic representations and reduce cross-task interference, while CDA recalibrates the shared classification head after each adapter merge via closed-form ridge regression. Extensive experiments across 19 generative domains demonstrate that DECODE achieves 99.36\% average accuracy with only 0.39\% forgetting, generalizes to eleven unseen generators with 95.36\% accuracy, and maintains robustness under common perturbations.

\bibliography{aaai2027}

\end{document}